\documentclass[conference]{IEEEtran}
\IEEEoverridecommandlockouts

\usepackage{cite}
\usepackage{amsmath,amssymb,amsfonts}
\usepackage{algorithmic}
\usepackage{graphicx}
\usepackage{textcomp}
\usepackage{xcolor}
\usepackage{url}
\usepackage{hyperref}
\usepackage{csquotes}
\usepackage{makecell}
\usepackage{subcaption}
\usepackage{caption}
\usepackage{balance}
\def\BibTeX{{\rm B\kern-.05em{\sc i\kern-.025em b}\kern-.08em
    T\kern-.1667em\lower.7ex\hbox{E}\kern-.125emX}}
\begin{document}

\title{HuNavSim 2.0: A Enhanced Human Navigation Simulator for Human-Aware Robot Navigation\\
\thanks{This work was partially supported by the Technology Exchange Programme of the EuROBIN project (GA 101070596) funded by the European Union, and the projects PICRAH4.0 (PLEC2023-010353) and NORDIC (TED2021-132476B-I00), funded by the Spanish Research Agency and the Ministry of Science (MCIN /AEI/10.13039/501100011033) and the European Union NextGenerationEU/PRTR}
}

\author{\IEEEauthorblockN{Miguel Escudero-Jim\'enez, No\'e P\'erez-Higueras, Andr\'es Martínez-Silva, Fernando Caballero, Luis Merino}
\IEEEauthorblockA{\textit{Service Robotics Lab} \\
\textit{University Pablo de Olavide.} Seville, Spain \\
\{mescjim, noeperez, amarsil1, fcaballero, lmercab\}@upo.es}
}

\maketitle

\begin{abstract}
This work presents a new iteration of the Human Navigation Simulator (\textit{HuNavSim}) \cite{perez23-hunavsim}, a novel open-source tool for the simulation of different human-agent navigation behaviors in scenarios with mobile robots. The tool, programmed under the ROS 2 framework, can be used together with different well-known robotics simulators such as Gazebo or NVidia Isaac Sim. The main goal is to facilitate the development and evaluation of human-aware robot navigation systems in simulation. In this new version, several features have been improved and new ones added, such as the extended set of actions and conditions that can be combined in Behavior Trees to compound complex and realistic human behaviors.   
\end{abstract}

\section{Introduction}

Today, mobile robots are moving from factories to domestic environments shared with humans. In such scenarios, human-aware navigation and interaction is becoming a relevant topic in robotics. 

The development of mobile social robots presents two primary challenges. First, conducting experiments involving real human participants is both costly and difficult to execute or replicate, except under highly controlled and limited conditions. Consequently, simulating realistic human navigation behavior becomes essential for the advancement of robot navigation techniques.  
Secondly, performance evaluation in a human-aware navigation context is another pending task in many simulation approaches. The evaluation must go beyond traditional performance metrics such as navigation efficiency, encompassing also the safety, legibility, and comfort of human co-inhabitants. However, some of these characteristics are inherently subjective and difficult to quantify mathematically, leading to a lack of consensus within the research community on the appropriate metrics for human-aware navigation \cite{Francis25_principles, Phani24_survey}.  

Most state-of-the-art simulation approaches employ crowd movement models to control the navigation of simulated human agents. Although such models are effective in reproducing global crowd behaviors, they often lack fidelity at the individual level. Specifically, these models typically assign uniform behavior patterns to all agents, leading to identical responses in similar scenarios regardless of environmental or contextual variations. As a result, the simulations fail to capture the diversity and unpredictability of real human navigation, particularly in the presence of robots. Moreover, human behavior is limited to navigation from one point to another in most cases. This lack of behavioral variability limits the utility of these simulations for developing and evaluating socially-aware robot navigation systems. In \textit{HuNavSim 2.0}, we employ noisy local navigation models to add variability and behavior trees to global and complex control of human actions.   

Furthermore, most current benchmarking tools introduce their own distinct set of evaluation criteria. Although the development of socially-relevant metrics remains an open research area, the lack of standardized metrics makes it difficult to perform fair comparisons between different social navigation approaches. In \textit{HuNavSim} we decided to collect the most relevant metrics found in the literature, as well as a flexible system to easily add new metrics.

\textit{HuNavSim 2.0} contributes to alleviate these issues by continuing the work initiated with \textit{HuNavSim 1.0} \cite{perez23-hunavsim}. The contributions of this new version are as follows.

\begin{figure}[!t] 
    \centerline{
    \includegraphics[scale=0.14]{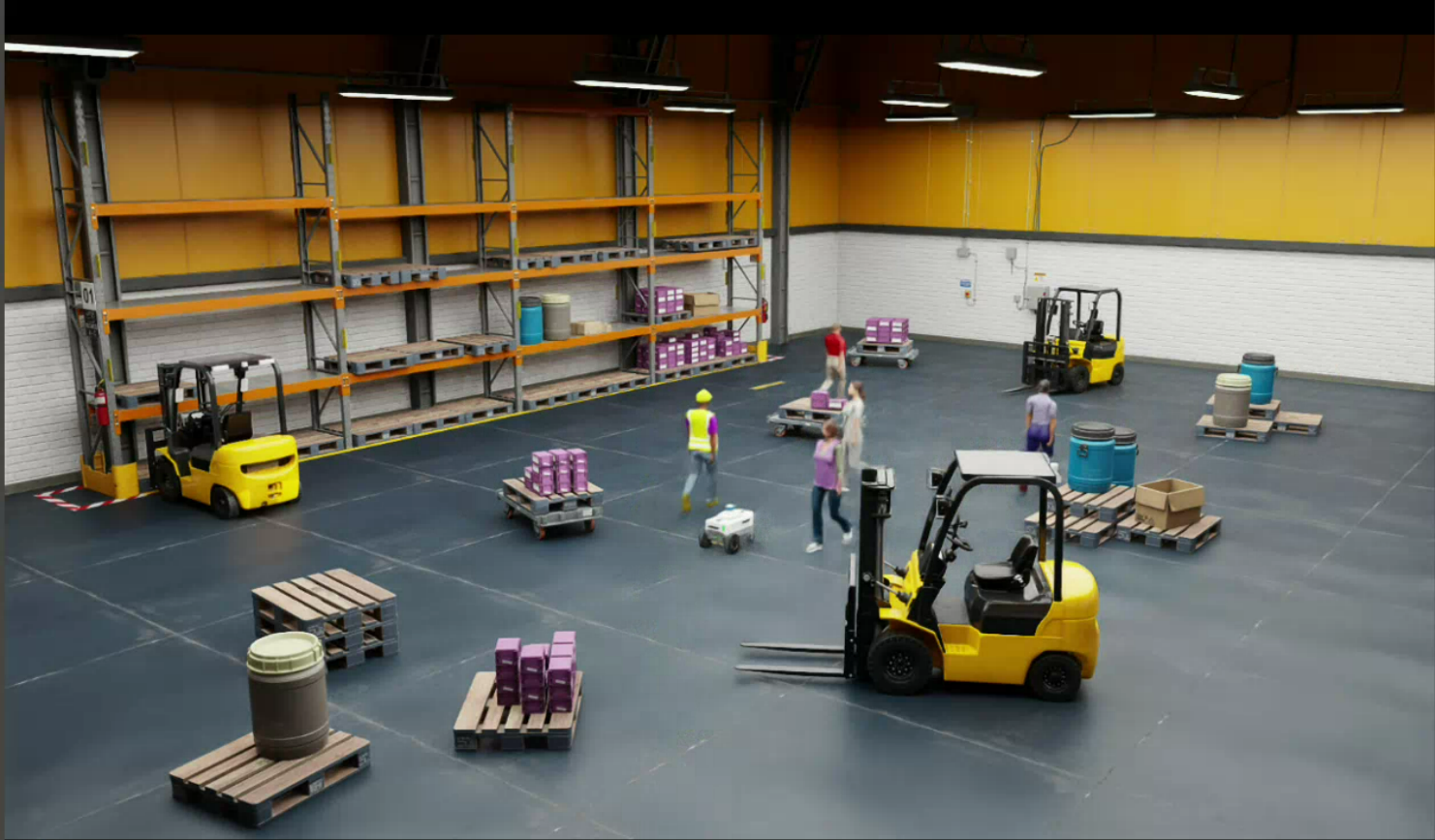}}
    \caption{Capture of HuNav agents in the NVidia Isaac Sim}
    \label{fig:isaac_people}
\end{figure}

\begin{itemize}
    \item[i)] A rich set of navigation actions and conditions to compound complex and realistic human navigation behaviors through
    Behavior Trees \cite{behavior_trees}. 
    \item[ii)] An enriched human local navigation by adding controlled noise to the parameters of the navigation model, leading to variable and more realistic human navigation.   
    \item[iii)] A set of different wrappers to use the tool along with well-known robotics simulator like Gazebo, Webots or NVidia Isaac Sim (see Fig. \ref{fig:isaac_people}). 
    \item[iv)] A new system to easily deploy the system and create new scenarios.
\end{itemize}

\section{Related work}

Several simulators and benchmarking tools have been developed to address human-aware navigation problems. This work is based on the previous tool \textit{HuNavSim} \cite{perez23-hunavsim}, which has been improved and extended. In this section, we briefly review existing approaches and differences with \textit{HuNavSim 2.0}.

The Social Force Model (\textit{SFM}) \cite{Helbing1995SocialFM} is a well-known model of human navigation through force-based interactions. Approaches like \textit{PedSimROS}\footnote{\url{https://github.com/srl-freiburg/pedsim_ros}} or \textit{MengeROS}\footnote{\url{https://github.com/ml-lab-cuny/menge_ros}} \cite{MengeROS_Aroor2017} integrate \textit{SFM} into the ROS ecosystem to simulate human navigation (unfortunately, deprecated versions of ROS). \textit{MengeROS} also offers different collision avoidance approaches such as \textit{ORCA} \cite{orca_2011} aor Pedestrian Velocity Obstacle (\textit{PedVO}) \cite{PedVO_2012}. In addition, it includes global path-planning algorithms. These approaches do not include an evaluation system or other features and are no longer maintained. \textit{ImHuS} (Intelligent Human Simulator) \cite{inhus_favier21a, imhus_Hauterville22} offers an alternative approach to simulating human behavior. It provides a limited set of individualized behaviors, but relies on ROS 1 and a social robot navigation system (\textit{HATEB2} \cite{hateb2_teja20}) to simulate human navigation, which can reduce realism compared to crowd-based models. In contrast, \textit{HuNavSim} uses modular and customizable Behavior Trees to model human behaviors. 

More ambitious and complete frameworks, such as \textit{CrowdBot}\footnote{\url{http://crowdbot.eu/CrowdBot-challenge/}} \cite{crowdbot_ICRA21} and \textit{ SEAN (Social Environment for Autonomous Navigation)} \footnote{\url{https://sean.interactive-machines.com/}} \cite{sean2_Tsoi_RAL22, sean_Tsoi20}, are built on the Unity game engine and ROS 1. These tools aimed to establish standard benchmarks for robot navigation in populated environments. In contrast, \textit{HuNavSim 2.0} provides greater flexibility, allowing integration with multiple simulation platforms and supporting a broader range of realistic human navigation behaviors. Unlike the fixed metric sets in \textit{CrowdBot3} and \textit{SEAN}, \textit{HuNavSim 2.0} offers an extensive, configurable, and easily extendable set of evaluation metrics.

More recent approaches, like \textit{SocialGym 2.0} \cite{holtz2022socialgym, socialgym2023} or \textit{Arena-ROSnav} \cite{KästnerRSS24_arena30} allows the training of human-aware navigation approaches based on (Deep) Reinforcement Learning. It uses the \textit{Pedsim} library to guide the movements of pedestrians. The version 4.0 of \textit{Arena} \cite{shcherbyna12024_arena40} runs on ROS 2 and integrates \textit{HuNavSim} (v1.0) to control the navigation of human agents. It uses large language models (LLMs) for the generation of scenarios (static environment) and a small and fixed set of metrics for evaluation. Previously, the work \cite{marpally2024socrates} also used \textit{HuNavSim} for human behavior and VLMs for scenario generation. \textit{SocialGym 2.0} also uses the Pedsim library to model pedestrian movement. However, it lacks support for individualized human behaviors and does not provide a dedicated set of metrics for evaluating human-aware navigation. Another approach that uses LLMs is \textit{Conav} \cite{li2024conav}. It is more focused on benchmarking collaborative tasks between human and robot than on human-aware navigation.

Lastly, benchmarking tools such as \textit{BARN (Benchmark for Autonomous Robot Navigation)}\footnote{\url{https://www.cs.utexas.edu/~attruong/metrics_dataset.html}} \cite{BARN_Perille2020} and \textit{Bench-MR}\footnote{\url{https://github.com/robot-motion/bench-mr}} \cite{bench-mr_21} target general navigation in cluttered environments. Although effective in evaluating obstacle avoidance and planning, they do not consider the social dynamics of shared human-robot spaces. For social navigation, \textit{SocNavBench}\footnote{\url{https://github.com/CMU-TBD/SocNavBench}} \cite{socnavbench_Biswas21} is a benchmark that uses replayed real-world pedestrian data from open datasets, allowing evaluation in realistic environments but lacking interactivity, since pedestrian trajectories remain unaffected by robot actions.
 
In summary, none of the previous tools offers the flexibility to create complex and realistic human navigation behaviors such as \textit{HuNavSim 2.0}. Furthermore, it is integrated into a wider variety of robotics simulators. Finally, neither of them includes the complete set of \textit{HuNavSim} metrics for the evaluation of social navigation. 
These features of \textit{HuNavSim 2.0} will be described in the following sections. 

\section{HuNavSim overview}

The basic functioning of the \textit{HuNavSim} tool is as follows. It has been devised to control the pose and behavior of human agents spawned within a base robotics simulator, such as Gazebo, Isaac Sim, or Webots, which also simulates the robot and the environment. Then, a communication wrapper is required to interface \textit{HuNavSim} with the chosen simulator.

\begin{figure}[!t] 
    \centerline{
    \includegraphics[scale=0.35]{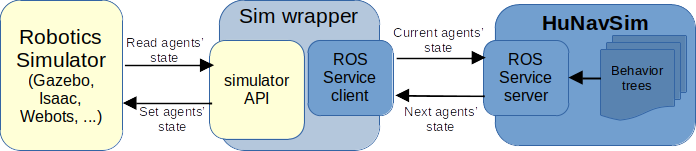}}
    \caption{Scheme of HuNavSim with Robotics simulators}
    \label{fig:diagram}
\end{figure}

In each simulation step, the wrapper sends the current state of the agents to the HuNavSim manager, which computes their next state based on the behavior trees and updates it in the base simulator (see Fig. \ref{fig:diagram}).

Another module is the evaluator, which logs experiment data and computes evaluation metrics at the end of the simulation. The results are saved as output files that contain the simulation data and the metric values.

A more detailed description of the core system can be found in our previous work \cite{perez23-hunavsim}. In version 2.0 of the tool, we have enhanced the capabilities to control and generate complex human navigation behaviors. We also have expanded the set of robotics simulators that can be used with \textit{HuNavSim} among other functionalities. These new features are explained in Sections \ref{sec:socialBeh} and \ref{sec:simulators}. 

The complete documentation and code of the \textit{HuNavSim 2.0} is available at \url{https://github.com/robotics-upo/hunav_sim} 

\begin{figure*}[!t] 
    \centerline{
    \includegraphics[scale=0.34]{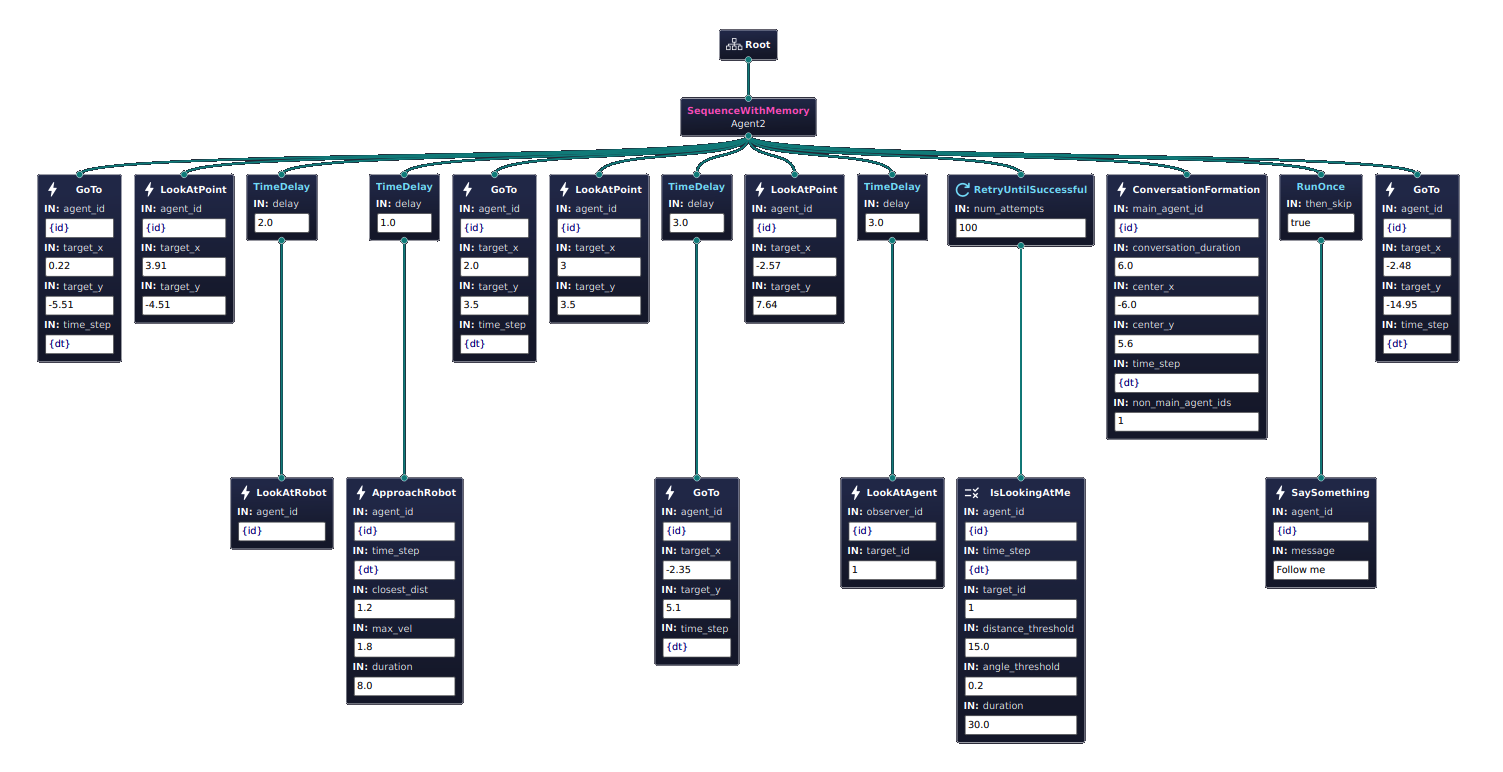}}
    \caption{Example Behavior Tree of the Worker 2}
    \label{fig:bt}
\end{figure*}

\section{Enhanced Social Human Behavior} \label{sec:socialBeh}

\subsection{Human local navigation model}

\textit{HuNavSim} employs the Social Force Model \cite{Helbing1995SocialFM} and its extension to groups \cite{Mussaid_ext_sfm09, Mussaid_ext_sfm10}, as the underlying model to lead the movement of human agents. This model, and many others, generally apply the same behavior patterns to all agents, resulting in uniform responses across similar scenarios. Consequently, they fail to capture the variability inherent in real human navigation.

In \textit{HuNavSim 2.0} we add behavior variability by introducing noise into the force factor parameters of the \textit{SFM}. We performed a parameter sensitivity analysis to identify feasible ranges that produce small yet realistic variations in human agents' navigation trajectories. Users can select from three configuration modes: (i) use default parameter values for consistent, repeatable behaviors; (ii) specify custom values tailored to specific applications; or (iii) enable stochastic behavior by sampling parameter values from normal distributions defined within the identified ranges, thereby introducing controlled variability into the simulation.

\subsection{Behavior Trees for human navigation behavior}

One powerful feature of \textit{HuNavSim} is the use of Behavior Trees to orchestrate the navigation behavior of human agents. In \textit{HuNavSim 1.0},
beyond navigation to indicated goals, human behavior was restricted to six possible reactions to the presence of a robot (\textit{regular}, \textit{impassive}, \textit{surprised}, \textit{curious}, \textit{scared}, and \textit{threatening}). 

Therefore, human agents can interact with the robot, but the interaction with each other is limited to walking together or avoiding other humans while walking. However, Behavior Trees are powerful tools for composing different tasks. We intend to exploit that fact with the aim of improving the overall interaction between humans and humans and between humans and
robots.

To achieve this goal, in \textit{HuNavSim 2.0} we implemented a set of different actions and conditions far beyond the navigation from one point to another and the previous reactions. These actions and conditions are programmed as Behavior Tree nodes and can be combined in a Behavior Tree to construct complex and realistic behaviors for human agents. The new set of conditions enables triggering actions according to the events that might occur. Thus, human agents can be aware of different social conditions that can vary or affect their behavior. The complete list of nodes and their characteristics can be consulted in the \textit{HuNavSim 2.0} Github repository previously indicated.
 
Next, we present the BT functionality with an example of two workers in a warehouse environment (scenario shown in Fig. \ref{fig:isaac_people}). The simulation video can also be watched on this \href{https://drive.google.com/file/d/1WlB_gY9iNMSHQSX5HYzyRnygWYDSCLi6/view?usp=sharing}{link}. The goal is to set a realistic scenario far beyond the previous version of unrealistic behavior navigation from one point to another in a loop.

\begin{figure*}[htbp] 
    \centering
    \includegraphics[height=2.9cm]{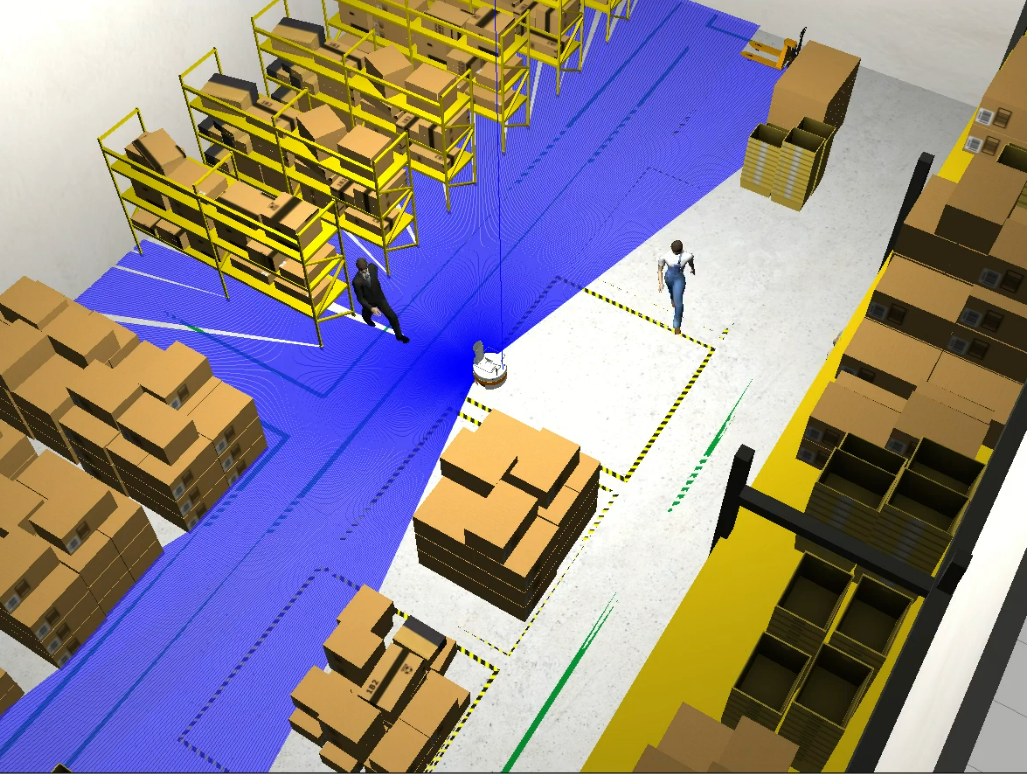}
    \includegraphics[height=2.9cm]{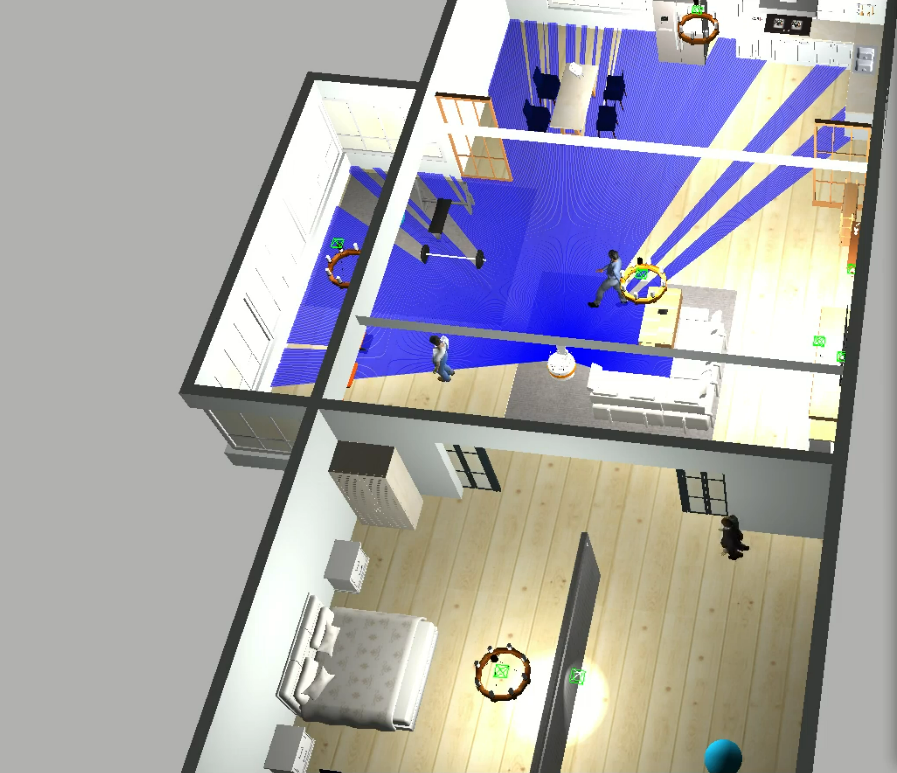}
    \includegraphics[height=2.9cm]{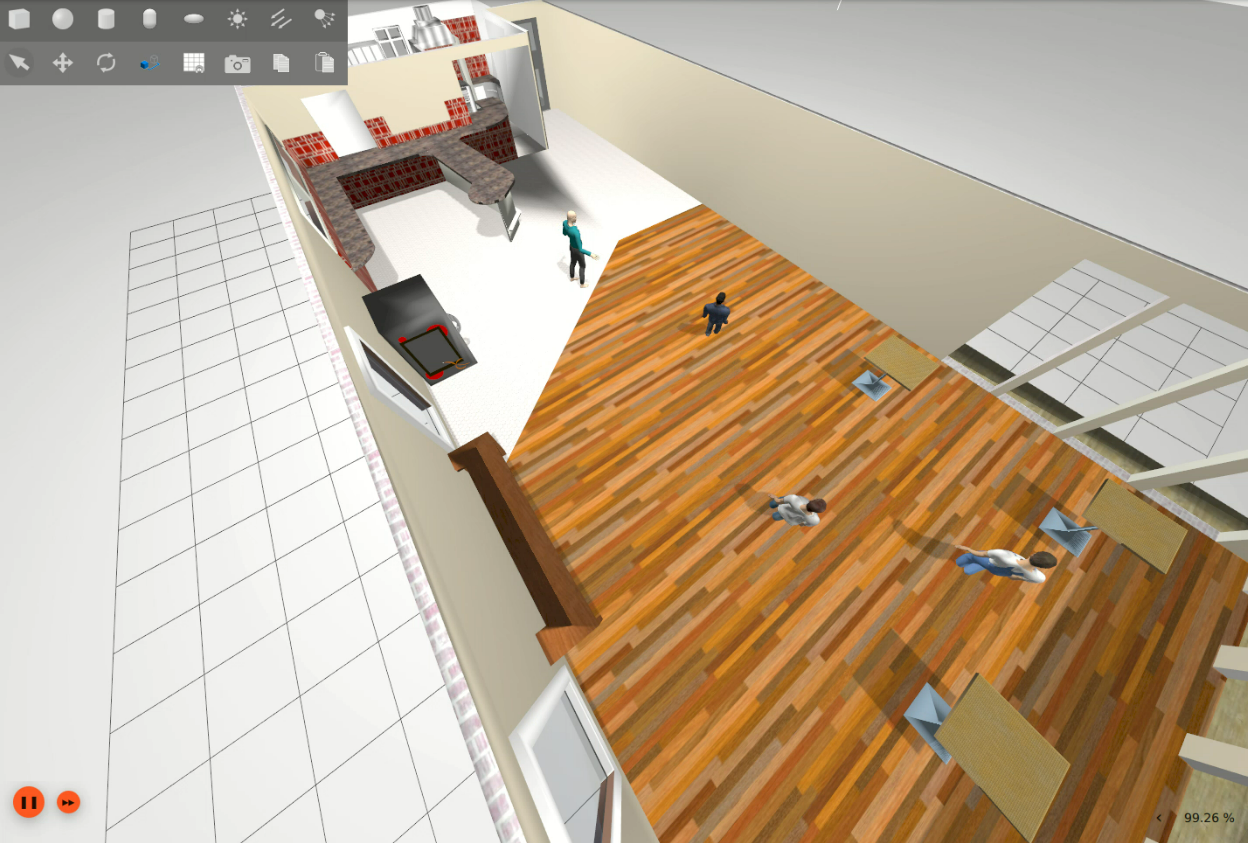}
    \includegraphics[height=2.9cm]{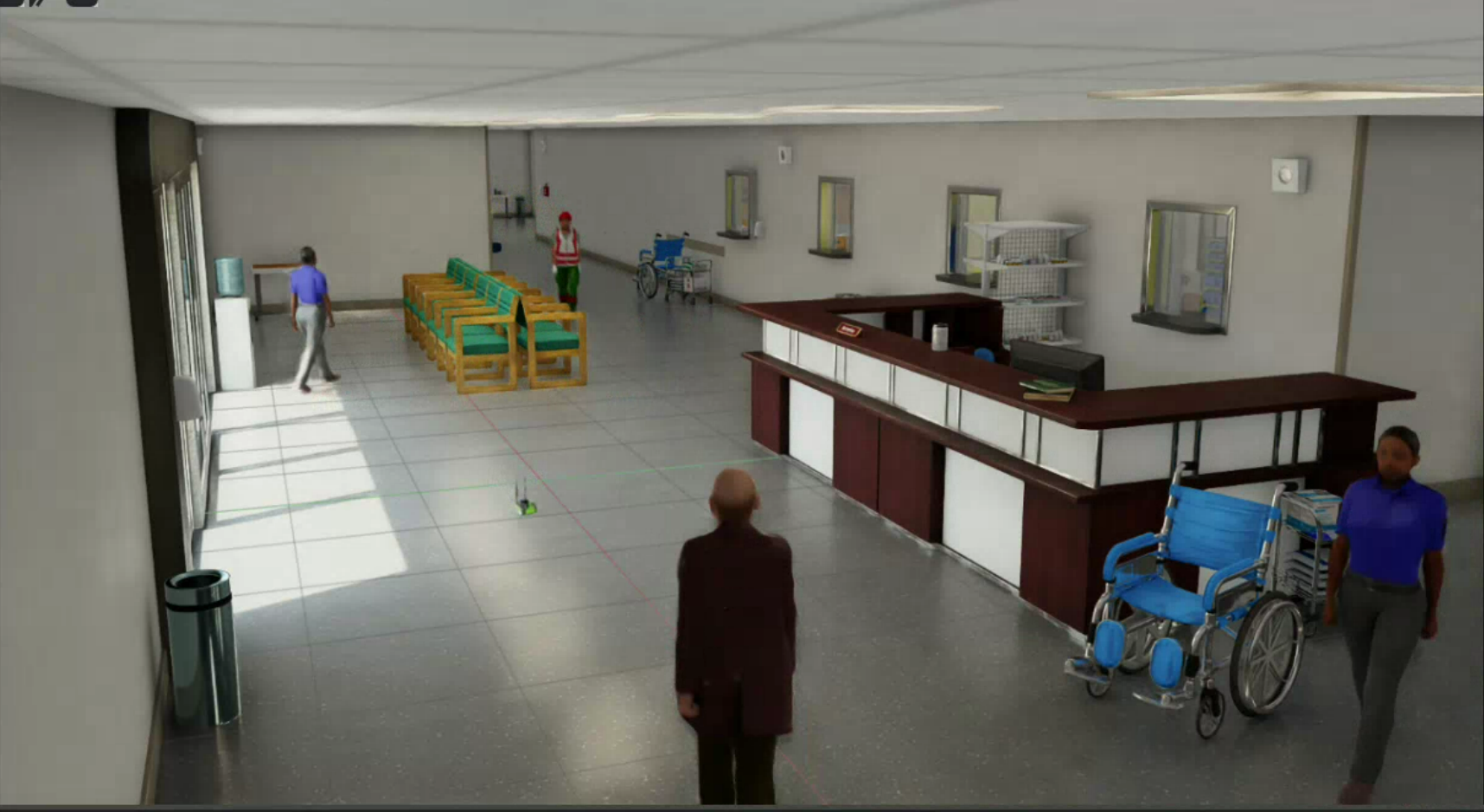}
    \includegraphics[height=2.9cm]{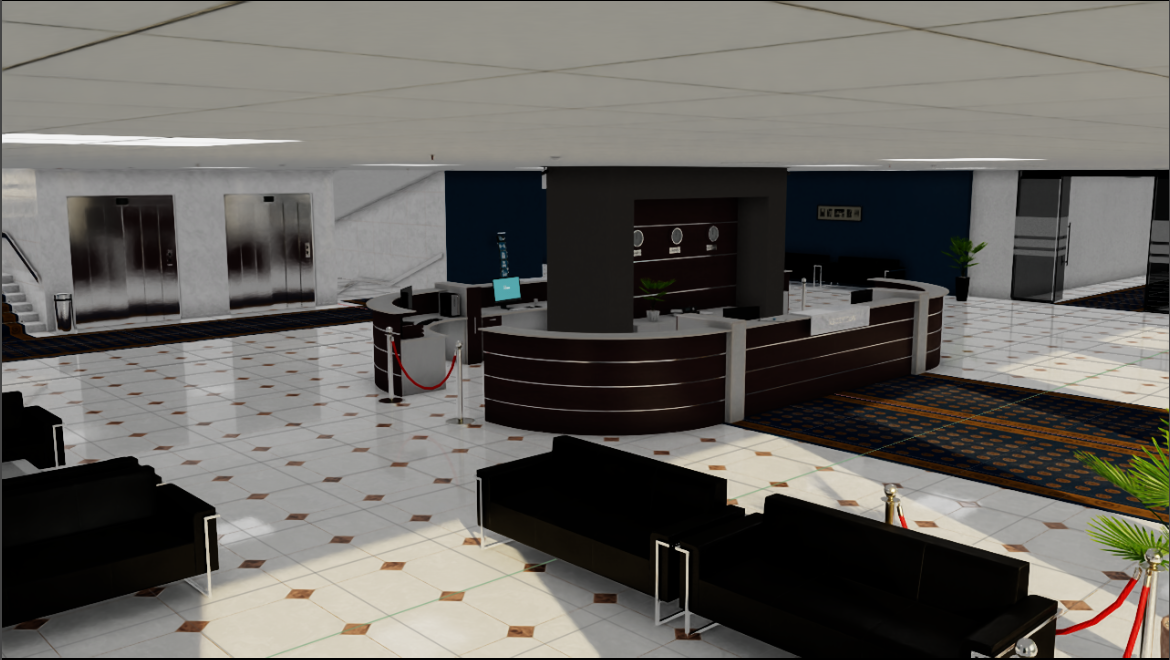}
    \includegraphics[height=2.9cm]{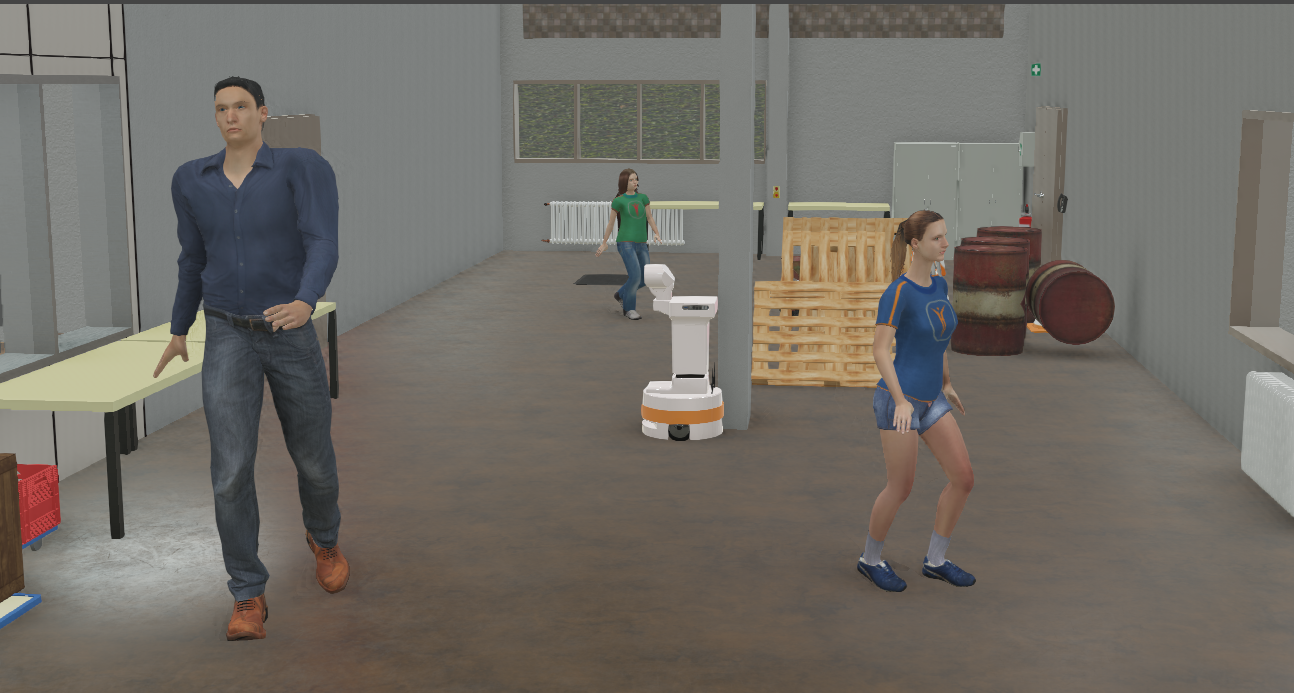}
    \includegraphics[height=2.9cm]{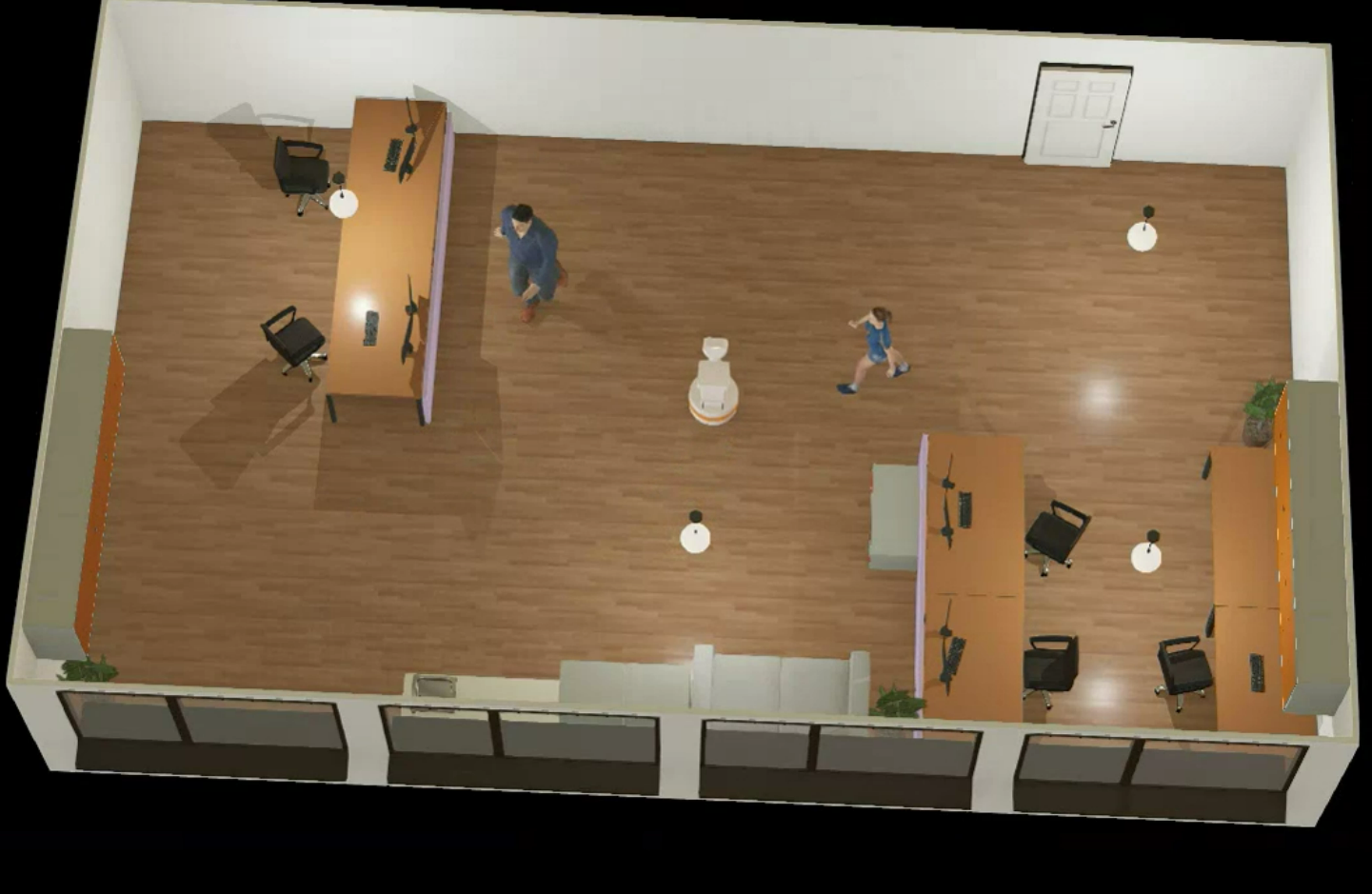}
    \captionsetup{justification=centering}
    \caption{Examples of simulators and some scenarios provided. From left to right and up to botton: Warehoure and home in Gazebo Classic, Cafe in Gazebo Fortress, Hospital and office hall in Isaac Sim, Industrial hall and office in Webots}
    \label{fig:scenarios}
\end{figure*}

In the example, the behavior of Worker 1 (white vest and red helmet) is
the following: he is checking some material in the warehouse. After that, he continues the round of checking material at different points in the scenario. This involves navigation to a waypoint (\textquote{\textit{GoTo}} node), looking at some point (\textquote{\textit{LookingAtPoint}}), and waiting some time (\textquote{\textit{StopAndWaitTimer}}). In his final location, the worker is checking whether Worker 2 is around (\textquote{\textit{isAtPosition}}). Once he has detected Worker 2, he looks at him (\textquote{\textit{LookAtAgent}}), and both move to form a conversation group initiated by Worker 2 (\textquote{\textit{ConversationFormation}}). The conversation is finished by Worker 2 who tells Worker 1 to follow him. Worker 1 executes \textquote{\textit{FollowAgent}} action when \textquote{\textit{isSpeaking}} condition is triggered. Worker 1 follows Worker 2 outside the scenario.

In parallel, Worker 2 (yellow vest and yellow helmet) starts his action, Fig. \ref{fig:bt} shows his BT. He begins checking the material at some location (\textquote{\textit{GoTo}} and \textquote{\textit{LookingAtPoint}}). Then, he notices the presence of the robot which captures his attention \textquote{\textit{LookingAtRobot}}. He moves close to the robot and looks at it for some time, simulating interest or assessment \textquote{\textit{ApproachRobot}}. Then, he moves toward a specific intermediate target and waits a bit. He repeats that behavior for the second and third targets, stopping at the location and looking to a specific direction. After that, he detected the presence of Worker 1 and looked at him \textquote{\textit{LookingAtAgent}}. Worker 2 checks whether Worker 1 is also looking at him (\textquote{\textit{IsLookingAtMe}} condition) and starts a conversation when the condition is fulfilled (\textquote{\textit{ConversationFormation}} action). After some time of conversation, Worker 2 says \textquote{\textit{follow me}} (using the \textquote{\textit{SaySomething}} action) and walks to a final target position, followed by Worker 1.

\section{Integration with Robot Simulators} \label{sec:simulators}

In contrast to \textit{HuNavSim 1.0}, which can be used only with Gazebo Classic simulator, \textit{HuNavSim 2.0} has considerably extended the set of integrated robotics simulators. Therefore, it can now be used to control the human agents spawned in four simulators. 

Table \ref{tab:simulators} shows the simulators available, as well as the examples of scenarios and robots provided for testing. Furthermore, the user can select from a different set of human skins for each base simulator. In case of the Gazebo Fortress simulator, we are working to extend the set of scenarios and provide a simulated robot. In any case, the user can use one of the provided examples or use his own base simulation. Captures of some of the provided scenarios can be seen in Fig. \ref{fig:scenarios}.  

\begin{table}[htbp]
\caption{Robotics simulators available for \textit{HuNavSim}}
\begin{center}
\begin{tabular}{c c c}
\hline
\textbf{Simulator} & \textbf{Example scenarios} & \textbf{Example Robots } \\
\hline
\textbf{Gazebo Classic} & Cafe, Warehouse, House & \thead{PAL PMB2\\(ROS Nav2)} \\
\hline
\textbf{Gazebo Fortress} & Cafe & \thead{\textit{Work in}\\\textit{Progress}} \\
\hline
\textbf{Nvidia Isaac Sim} & \thead{Warehouse, Hospital,\\Office hall} & \thead{Jetbot, Create3, \\ Carter (only this latter\\with ROS Nav2)} \\
\hline
\textbf{Webots} & \thead{Factory, Office,\\Industrial hall} & \thead{PAL TIAGo\\(ROS Nav2)} \\
\hline
\end{tabular}
\label{tab:simulators}
\end{center}
\end{table}

\textit{HuNavSim 2.0} is also distributed using Docker containers and a simple \textit{CLI} to guide the user through an easy installation process of the chosen system and then to run the examples or add new simulations or scenarios.

\subsection{Generation of new scenarios}

\begin{figure}[htbp]
  \centering
  \begin{subfigure}{0.53\linewidth}
    \includegraphics[height=9.05cm]{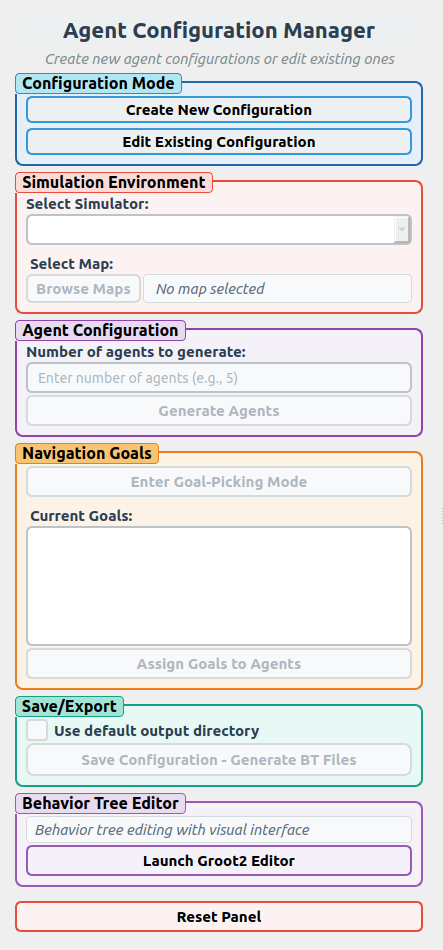}
    \caption{HuNavPanel: agent configuration.}
    \label{fig:actor_panel}
  \end{subfigure}
  \hfill
  \begin{subfigure}{0.40\linewidth}
    \includegraphics[height=9.05cm]{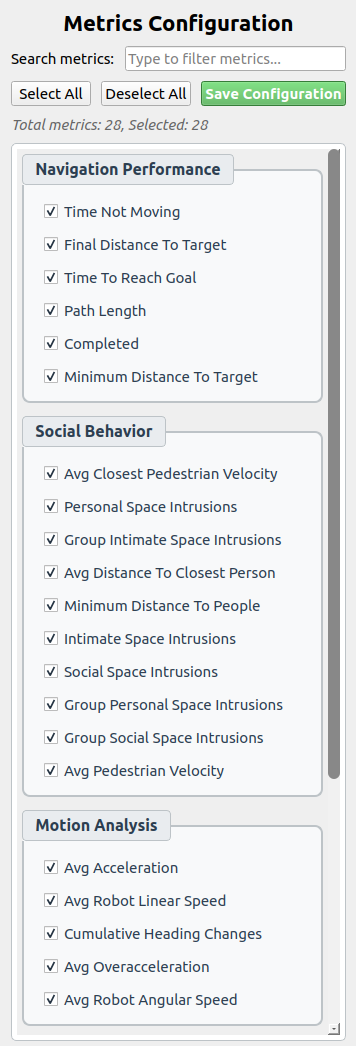}
    \caption{HuNavMetricsPanel: metrics selection.}
    \label{fig:metrics_panel}
  \end{subfigure}
  \caption{HuNav RViz~2 panel suite providing agent‑configuration and metrics‑selection functionality.}
  \label{fig:hunav_panels}
\end{figure}

\textit{HuNavSim~2.0} introduces an enhanced \emph{what-you-see-is-what-you-simulate} workflow, allowing complete scenario creation directly within RViz~2. Central to this are two dedicated Qt-based panels for agent configuration and metric selection (Fig.~\ref{fig:hunav_panels}), which, together with the simulation wrappers (Sec.~V), enable rapid progression from map setup to behaviour-tree synthesis.

\paragraph{Launch and initial set‑up} From the CLI the user can launch the creation of a new scenario. RViz~2 will be open with all HuNav panels. The \textit{Simulator\,\&\,Map} section lets the user select the target back‑end—Gazebo~Classic, Gazebo~Fortress, Isaac~Sim, or Webots—and load any 2‑D occupancy map supported by \texttt{nav2\_map\_server}.

\begin{figure}[htbp]
    \centering
    \includegraphics[height=9.05cm]{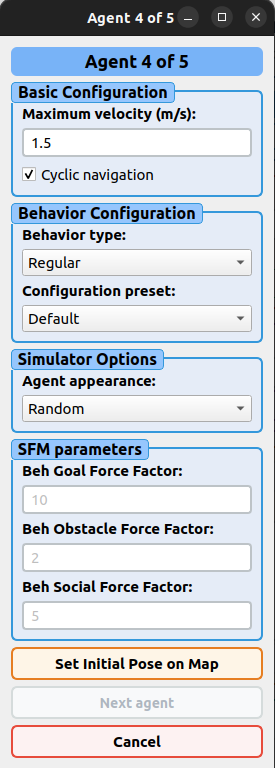}
    \caption{Sequential agent dialog}
    \label{fig:add_agent_dialog}
\end{figure}

\paragraph{Interactive agent creation}
Agents are configured sequentially through a dialog that combines parameter input with in-scene placement (Fig.~\ref{fig:add_agent_dialog}). For each agent, the user sets core motion limits, selects a behavior profile (e.g., regular, impassive), and adjusts Social-Force parameters—either keeping defaults, sampling, or fine-tuning them. A simulator-specific appearance catalogue ensures visual diversity. Clicking \emph{Set Initial Pose on Map} launches a click-and-drag tool to place and orient the agent marker. The panel then advances to the next agent until the population is complete.

\paragraph{Goal definition and assignment}
After placing all agents, the user can activate the \textit{Goal‑picking mode}, which allows them to select target locations by clicking on the map. Each selected point is represented as a coloured marker and added to a goal list. Once the desired goals are defined, the \emph{Assign Goals to Agents} dialog provides a dual-list interface to associate specific goals with individual agents. As goals are assigned, the corresponding navigation path is immediately visualized in the environment.

\paragraph{File export and BT synthesis}
Saving writes the full scenario to \texttt{[map]\_agents\_\textless{}...\textgreater{}.yaml} and auto‑generates one XML behaviour tree per agent (\texttt{\textless{}yaml-basename\textgreater{}\_\_agent\_\textless{}id\textgreater{}\_bt.xml}). Both files are saved to the wrapper's share directory. A shortcut opens Groot~2 for graphical inspection or refinement of the generated trees.

\paragraph{Re‑editing existing scenarios}
Opening an existing YAML repopulates RViz~2 with every agent, goal, and path, ready for point‑and‑click editing. Agents can be cycled and parameters and goals tweaked just as during creation, enabling rapid iteration and focused ablation tests.

\medskip
In short, the HuNav~RViz~2 panel turns scenario creation into an interactive, visual task while preserving YAML/BT flexibility and real‑time feedback. Further details and examples are available in the \href{https://github.com/robotics-upo/hunav_rviz2_panel}{\texttt{hunav\_rviz2\_panel}} repository.

\subsection{Evaluation and metrics}

The appropriate evaluation of human-aware navigation systems is still an open question in the navigation field. There is no common agreement in the community about specific benchmark scenarios and a suitable set of metrics \cite{Francis25_principles,Phani24_survey}. For these reasons, we keep \textit{HuNavSim} as an open system in which the user can modify/add scenarios and metrics in addition to the provided ones. 

Considering the scenarios, we provide examples in which common situations of crossing, overtaking, or passing in different directions are present. Furthermore, \textit{HuNavSim} includes a set of 28 metrics, which constitutes, to the best of our knowledge, the most comprehensive collection of metrics for human-aware navigation to date. The complete list is presented in our previous work \cite{perez23-hunavsim}. Detailed descriptions and corresponding mathematical formulations for each metric are available in the HuNavEvaluator section of the GitHub repository. The user can easily select which of these metrics they want to compute, and can add new ones easily. 

In \textit{HuNavSim 2.0}, we maintain the same system, just improving the way metrics can be calculated and adding two new interesting metrics. We have extended the evaluation system so that the user can decide when to start calculating metrics and when to stop it by calling ROS services. 

The new four metrics, called \textquote{danger and surprise metrics}, are obtained from the work of Singamaneni et al. \cite{Phani23_roman}. That gives a total of 32 metrics that can be obtained. 

\section{Conclusions and future work}

In this work, we have presented the new version of \textit{HuNavSim} (\textit{2.0}), an open-source software library designed to simulate human navigation behaviors in environments shared with robots. Developed in the ROS 2 framework, HuNavSim can control human agents across multiple general-purpose robotic simulators with a dedicated wrapper provided.

One of the key innovations of \textit{HuNavSim 2.0} is its implementation of human behaviors using Behavior Trees. We have programmed a complete set of actions and events that can be used to compound rich human behaviors. Behavior Trees which offer a modular and easily extensible structure to build complex and more realistic human navigation behaviors. In addition, the tool includes a comprehensive and flexible set of evaluation metrics for human-aware robot navigation compiled from the existing literature.

We also have improved the realism of the local navigation model by integrating controlled noise into the parameters of the Social Force Model. This gives us a variability in the navigation trajectories closer to real human navigation. 

Future work will focus on facilitating the creation of scenarios and behavior trees for human behaviors. The use of LLM or VLM will be explored.
We will also try to improve the underlying navigation model by adding other models in addition to classical SFM. Finally, we will closely follow the advances in human-aware navigation benchmarking to adapt the tool to the community needs.

\balance
\bibliographystyle{IEEEtran} 
\bibliography{hunav}

\end{document}